# MLOps Spanning Whole Machine Learning Life Cycle: A Survey

Fang Zhengxin[1], Yuan Yi[2], Zhang Jingyu[4], Liu Yue[4], Mu Yuechen[5], Lu Qinghua[4], Xu Xiwei[4], Wang Jeff[4], Wang Chen[4], Zhang Shuai[3] and Chen Shiping[1,4*]

[4]CSIRO, Data61, Sydney, Australia.
[1]Victoria University of Wellington, Wellington, New Zealand.
[2]University of Surrey, Surrey, United Kingdom.
[3]Amazon, California, USA.
[4*]CSIRO, Data61, Sydney, Australia.
[5]The University of New South Wales, Sydney, Australia.

*Corresponding author(s). E-mail(s): shiping.chen@csiro.au;
Contributing authors: timmyfangzx@gmail.com;
yy01071@surrey.ac.uk; jingyu.zhang@data61.csiro.au;
yue.liu@data61.csiro.au; muyueche@msu.edu;
dr.qinghua.lu@gmail.com; xiwei.xu@data61.csiro.au;
jeff.wang@data61.csiro.au; Chen.Wang@data61.csiro.au;
cheungshuai@outlook.com;

## Abstract

Google AlphaGo's win has significantly motivated and sped up machine learning (ML) research and development, which led to tremendous ML technical advances and wider adoptions in various domains (e.g., Finance, Health, Defense, and Education). These advances have resulted in numerous new concepts and technologies, which are too many for people to catch up to and even make them confused, especially for newcomers to the ML area. This paper is aimed to present a clear picture of the state- of-the-art of the existing ML technologies with a comprehensive survey. We lay out this survey by viewing ML as a MLOps (ML Operations) process, where the key concepts and activities are collected and elabo- rated with representative works and surveys. We hope that this paper can serve as a quick reference manual (a survey of surveys) for newcomers (e.g., researchers, practitioners) of ML to get an overview of the MLOps process, as well as a good understanding of the key technologies used in each step of the ML process, and know where to find more details.

**Keywords:** machine learning, MLOps, datasets, neural networks, gaze detection, text tagging

# 1 Introduction

Machine learning (ML) is the study of computer algorithms that can improve automatically through experience and data. In the last decades, the advance in computer hardware (CPU, GPU, ROM, Disk) and the availability of large amounts of various data have fostered ML research making it mature and robust. Google AlphaGo's win is one of the remarkable milestones indicating that ML was ready to be applied in real applications. Nowadays, ML has been widely used in various applications, from image recognition to automatic speech translation, from unmanned vehmedical medicine diagnosis, and from email filtering to terrorist attacks detection/warning [1].

ML is a board research area with many research branches and terminologies to address different problems such as classification, regression, clustering, dimensionality reduction and decision-making, deep learning (DL) [2], supervised learning [248], unsupervised learning [4], reinforcement learning [5]. Good survey papers are expected to collect, summarize, and explain these terminologies, the related technologies and their relationships or linkages. There have been numerous survey papers in ML literature. For example, [6] is an early survey paper on image registration techniques; [247] provides an overview for deep learning; [7] focuses on a review of applying deep learning to biometrics; [5] delivers a survey for reinforcement learning, and [8] and deep reinforcement learning. While these survey papers are helpful by summarizing the state-of-the-art on specific ML topics, there is still a lack of a global view of the ML field and the linkages among ML terminologies.

In addition, as need more data for training ML models and deploying the ML models in real applications, people are facing new challenges and requirements for ML research, such as the overhead of transferring massive data, data privacy, cybersecurity, Responsible-AI (RAI) and AI Ethics. While researchers are actively working on addressing these challenges and there are a few surveys to cover some of the thesis topics [42–44, 44–46]. However, little effort has been put to integrate these new concepts and technologies into the big ML puzzle.

This paper aims to present a clear picture of the existing ML technologies with a comprehensive survey. This survey is based on a taxonomy of viewing ML as a process of ML operations (MLOps). In each step in the MLOps process, the key concepts and activities are collected and elaborated with the well-selected representative work and the corresponding surveys. This paper can serve as a quick reference for researchers (a survey of surveys) and/or someone new to ML to get an overview of the whole ML process, a good understanding of the key activities conducted and technologies used in each step of the ML process, and know where to find more in details.

The rest of this paper is organized as follows. In the next section, we provide a brief introduction of our MLOps process model and the basic concept of each step. We then elaborate on each step in detail in Sections 3-10. Section 11 concludes this paper.

## 2  MLOps Process Model

Although there are many machine learning methods and algorithms for different purposes, they usually consist of several common activities / actions, which form an ML process. There have been considerable efforts to name these activities and group them to form one step in the ML process [48–51]. Still, there are no standard notations and models for these activities/groups and processes.

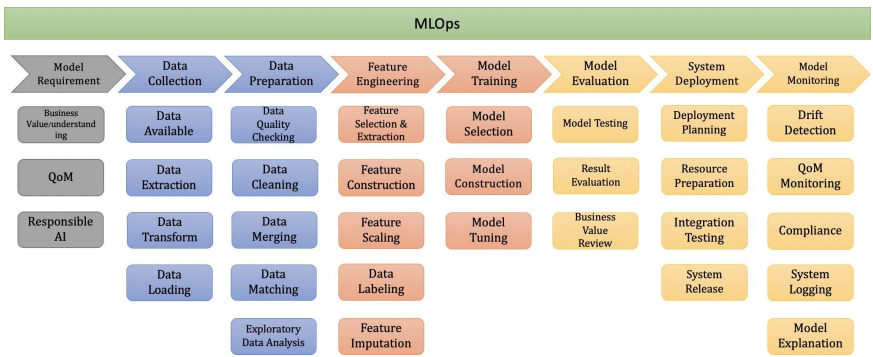

**Fig. 1**  The machine learning process.

In this paper, we view a machine learning process as *MLOps* by reconciling both existing literatures [48–50] and industrial best practices [52, 53, 55]. As shown in Figure.1, our MLOps-based model consists of eight steps, each of which includes a set of activities/tasks including:

1) **Model Requirement**: This is the first step for a machine learning pro- cess with a focus on the objectives and requirements for machine learning projects. The requirements cover: (a) Business Value (BV) - Why do this project, What are the core functionalities of the ML model; (b) Quality of Model (QoM) - How good the model is going to be; (c) Responsible AI - It is about how a specific organization is addressing the challenges around AI techniques from both an ethical and legal point of view.

2) **Data Collection**: Data is the source to build and refine ML models. As a result, data collection is an immediate step to identify: (a) what data is required; (b) if the required data is available; (c) how to get the available required data; (d) is the data format correct? If not, the data need to be transformed before useing in the following steps; (e) the training data needs to be labelled for supervised machine learning.

3) **Data Preparation**: The performance of machine learning model is acutely sensitive to the dataset. Data preparation consists of a series of preprocessing methods on the raw dataset. This step covers: (a) Data Quality Checking – define the data quality problem; (b) Data Cleaning - fix the errors in the raw dataset; (c) Data Merging – merge various datasets into one; (d) Data Matching - find records that refer to the same entity; (e) Exploratory Data Analysis (EDA) - analyze datasets to summarize their main characteristics.

4) **Feature Engineering**: This step transforms the raw feature space into features that can be used in model training. It can produce new features that improve model performance: (a) transforming the raw feature by feature selection, extraction, construction, and scaling; (b) label the target for unlabelled data; (c) imputation feature to deal with the missing values; (d) converting data to a specific format.

5) **Model Training**: There are four main types of models: supervised learning, unsupervised learning, semi-supervised learning, and reinforcement learning. To get the most suitable model for a dataset, the training step includes: (a) Model Selection – the criteria of selecting models; (b) Model Construction – architecture of model; (c) Model Tuning - adjusting the hyperparameters of the model to get better performance model.

6) **Model Evaluation**: After model training, we need to evaluate the model to validate how well it performs on an unknown dataset. There are several metrics for evaluating machine learning products that answer questions like how well is this model doing; how it compares to other existing technologies with the same problem. As illustrated in Figure.1, model evaluation typically begins with (a) model testing; after that, (b) the result evaluation task; then (c) business review.

7) **Model Deployment**: Model deployment involves all of the steps, processes, and activities to make machine learning models ready for use. In this section, we discuss the workflows that enable efficient and effective implementations of a machine learning model and when to release it to end-users. A model deployment process includes deployment planning, resource preparation, integration test, and system release.

8) **Model Monitoring**: The performance of a deployed machine learning model may degrade over time in a production environment. The goal of model monitoring is to actively track the model matrix and identify potential problems before it impacts the business negatively. There are five categories of model monitoring: (a) Drift Detection; (b) QoM Monitoring; (c) Compliance; (d) System Logging; (e) Model Explanation.

The above ML process model will be used as a reference ML process model to guide our survey. For each step, we apply the same methodology and procedure to conduct our survey as specified as follows: 1) For each task in each step, at least five papers are selected and used for reviewing and comparing the relevant methods and technologies; 2) Our survey paper selection criteria are specified as follows: (a) half of the papers are selected from well-known relevant venues based on numbers of citations; and (b) the other half of the papers are selected by taking the following aspects into account, i.e. freshness (publication year) and coverage to avoid duplication; 3) In addition to academic publications, we include widely adopted existing technologies and tools to reflect the best practice for a specific task/topic; 4) At the end of each survey for a specific topic, we use a table to summarise and compare the surveyed technologies for easy reading and reference.

## 3 Model Requirement

Model requirement is significant as it identifies the right problems that the machine learning project aims to solve. However, identifying requirements can be challenging as existing requirement engineering methods are not directly applicable [86]. There is increasing research underlining the need for appropriate model requirements in machine learning projects [87]. As there is a lack of commonly recognised requirement methodology for machine learning, model requirement should be conducted case by case and hereby, we list four fundamental dimensions that should be contemplated when analysing the requirements of a machine learning model, including business value, quality of model, human value, and ethics.

First, stakeholders should precisely delineate the objectives and according to functionalities of the model regarding the overall **business values**, subsequent process and extensive business ecosystem. Considering the following question may help identify the requirements, including ML models, training data, and system autonomy: 1) What are the potential business values and problems? 2) What tools can help identify and present the values and problems? 3) Which values and problems should be prioritised? 4) Who are involved when solving the problems? 5) What functions are needed to solve the problems? Specifically, Ahmad et al. [88] study the modelling languages and requirements notations that can be utilised in AI systems, including goal-oriented requirement engineering, UML, signal temporal logic, traffic sequence charts, and conceptual model.

The next step is to analyse the quality of the model and whether the model can meet both **functional requirements** that embody the model features and core business rules such as functional completeness, correctness, and appropriateness, and also **non-functional requirements** including performance efficiency, usability, adaptability, etc. Evaluating the quality of the model can refer to the selected tools mentioned in the previous step. Functional requirements can be quantified based on the model's ability to describe and resolve the business values and problems respectively, whereas non-functional requirements can be measured regarding various software quality attributes [78, 89]. Specifically, Gruber et al. [90] apply SysML to visualise both kinds of requirements and to explore the interrelationships.

**Quality of Model (QoM)** is usually used to evaluate how good the model is. It will affect the subsequent procedures of the model, so it is necessary to pay attention to and improve the quality of the model. When analysing the quality of the model, researchers consider the following factors.

1) **Accuracy**: Accuracy reflects the relationship between the target of the model setting and the result of the model. It is the key to judging the quality of a model, which provides the researcher with the most intuitive way to observe whether the results of a model meet the needs and expectations. All the information shown in the model needs to be clear and explicit. All the algorithms should be based on logic and reasoning, and balance the correlation among different ways. In addition, the model can be able to minimized human error within the controllable range. Although the systematic error is allowed, a good machine learning model should be able to minimise it and get the conclusion in accordance with its purpose. A survey [147] paper mentioned the way of using the data validation and cleaning techniques after collecting data to improve the accuracy of deep learning.

2) **Robustness**: Robustness is used to test the tolerance and anti-interference ability of a model or system. Under the condition that the initial settings are unchanged, a good model should be able to keep the original state in the face of internal or external perturbations. Improving the robustness of the model is beneficial to maintaining the integrity of the model and improving the stability of the model. Defensive distillation is one of the original approaches to improving machine learning robustness. However, Carlini[145] supposes that the high-confidence adversarial still creates a threat to a system. According to [152], the new classification of malicious attacks and defences will help the optimisation problem of robustness.

3) **Generalisation ability**: A good model should be work well in the goodness of fit, and overfitting and underfitting should be avoided. Overfitting can make the program of the model overly strict for a particular assumption, resulting in the inability to support a large amount of data and the deviation in the prediction results. Underfitting can make the program of the model lose the ability to determine the characteristics of the data so that the model's data of the model cannot be calculated and predicted well[146].

4) **Simplicity**: When the quality of the model is evaluated, it should be simplified as much as possible. Once all conditions are met, no additional factors should be added to the original model to make it complicated. Moreover, the same information should not be described repeatedly. The model should sift through existing algorithms and select results that are more in line with mathematical expectations. At the same time, researchers should compare and evaluate the strengths and weaknesses of different algorithms to choose the most suitable method for the model. In addition, when conditions permit, overly complicated operations should be avoided during the model's deployment process to reduce the burden of the model on users [153].

5) **Consistency**: The goal and information of the model should remain the same to avoid the problem of inconsistency. There should be no logical conflicts and contradictions in the model itself, and all information should maintain unity and harmony with the final purpose.

Further, model requirements can be more inclusive to enhance the trustworthiness of related code, algorithms, data, etc. In particular, **human values** should be acknowledged when identifying the model requirements, as specific values are related to software quality attributes. Note that there might be trade-offs between certain human values, as well as the software quality attributes discussed above. Embedding human values in machine learning has become an important field of study, to reflect the impacts of the machine learning project on society and people [79]. Some researchers have focused on human values and their operationalisation in machine learning. For instance, architecture and process decisions as well as design patterns in the analysis of model requirements can improve certain human values such as privacy, fairness, security, and accountability [82, 84, 203].

In addition, this step can move forward to the respect of broader **legal compliance** and **ethical responsibilities**. Legal compliance represents the minimum standards of human behavior, while ethical responsibilities denote the maximum standards of human behavior [91]. There are already many high-level ethical principle frameworks [83, 92] that help comprehend significant ethical and important legal principles. Specifically, identifying ethical responsibilities depends on where the models are planned to deploy, and how stakeholders (e.g., regulators) design related governance principles and standards [100, 101]. An example would be the Australia's Artificial Intelligence Ethics Framework[1], which can be adopted to identify the requirements in a machine learning project.

1) **Human, social and environmental wellbeing**: The ML model should be beneficial to individuals, society and the environment throughout its lifecycle.

2) **Human-centred values**: The ML model should respect human rights, diversity, and the autonomy of individuals throughout its lifecycle.

3) **Fairness**: The ML model should be inclusive and accessible, and should not involve or result in unfair discrimination against individuals, communities or groups throughout its lifecycle.

4) **Privacy protection and security**: The ML model should respect and uphold privacy rights and data protection, and ensure the data security throughout its lifecycle.

---

[1]https://www.industry.gov.au/data-and-publications/australias-artificial-intelligence-ethics-framework/australias-ai-ethics-principles

5) **Reliability and safety**: The ML model should reliably operate in accordance with the intended purpose throughout its lifecycle.
6) **Transparency and explainability**: There should be transparency and responsible disclosure so stakeholders can understand when they are being significantly impacted by the ML model, and can find out when the model is engaging with them.
7) **Contestability**: When the ML model significantly impacts a person, group or environment, there should be a timely process to allow relevant stakeholders to challenge the use or outcomes of the model.
8) **Accountability**: Those responsible for the different phases of the ML model lifecycle should be identifiable and accountable for the outcomes of the model, and human oversight of the model should be enabled.

## 4  Data Collection

Data collection has recently become a critical issue in machine learning for the following reasons. Firstly, as machine learning is more widely adopted in various application domains, it is always the case that there is a lack of training data. Some mature areas of machine learning, like objection detection, and machine translation benefit from large volumes of training data collected for decades. Moreover, deep learning is an area with a high demand for large amounts of data, because deep learning can generate features automatically, which will save a lot of time in data preparation; however, on the other hand, deep learning should require more significant amounts of data to achieve a good performance. As a result, there is a crucial need for data collection methods for the machine learning domains, which will be discussed by dividing it into two parts: 1) data acquisition; 2) data extract, transform, and load (ETL).

### 4.1  Data Acquisition

Data acquisition is about acquiring data for specific ML tasks. A survey on data collection for machine learning [19] presents three approaches to data acquisition: data discovery, data augmentation, and data generation.

Data discovery is a variety of approaches for sharing generated datasets and searching for suitable datasets. Data sharing is about data systems designed to make the generated data easy to share. These systems may focus on collaborative analysis, publishing on the Web, or both. Kaggle is a collaborative web-based system that makes sharing datasets easy by hosting competitions. As for data searching, it is data systems mainly designed for searching datasets. The data lake is a concept of a centralized repository containing large amounts of data in raw format that is ready for authorized access at any time. Data searching systems have become more popular with the advent of data lakes. Google Data Search (GOODS) [20] is a system that extracts metadata ranging from basic information (owners, timestamps, schema and so on) to relationships among datasets (similarity and provenance). And then users can use

keyword queries on the GOODS to find datasets, view profiles of each dataset, and track the provenance of a dataset to see those datasets that it relies on.

Data augmentation is another approach to acquiring data by augmenting existing data with external data. Deriving latent semantics, entity augmentation and data integration are the existing approaches for machine learning tasks to extend the existing dataset. Firstly, a popular is generating and using embeddings to represent words, entities, or knowledge. For example, Word2vec [22] is an influential work that generates word vectors for given words, which have been widely used in natural language processing (NLP). Secondly, entity augmentation [23] is performed by filling in missing values in entities by matching multiple Web tables using schema matching. As for data integration, it's an approach to extending existing acquired from the other data sources.

If a machine learning task has no training datasets at all then an option is to generate the datasets is manually or automatically. Crowdsourcing is a method used for generating data manually, which can be divided into two steps: gathering data and preprocessing data. Generating synthetic data is increasingly being used in machine learning due to its low cost. Generating data from probability distribution by using tools like Scikit-Learning is [24] a simple method for generating synthetic data. In addition, Generative Adversarial Networks (GANs) [25] is a deep learning method that can automatically generate data.

## 4.2  Data ETL

The ETL (Extraction, Transform and Load) is a process of integrating multiple data sources into a single, consistent, safe place, e.g. data warehouses.

1) **Extraction:** It's a process of extracting data from different sources and formats like a relational database, XML, CSV files, other etc. The extracted data are saved in a staging area before being stored into a data warehouse so that those data have an opportunity to validate the data. The error data found are sent back to the source system to correct the wrong values during the validation process.
2) **Transform:** In this process, the extracted data are properly transformed/-formatted as required before loading into the data warehouse. The common tasks of this step are: filtering data, converting codes, transforming data formats, etc.
3) **Load:** In the load process, the transformed data is moved from the staging area to the data warehouse.

[27] focuses on the definition of ETL activities and proposes various entities that can capture the semantics of the ETL process. Also, it uses an example to show how these entities can be used to present the conceptual model. It abstractly describes a framework for ETL processes. In its framework, the data from various data sources (e.g. typically, relational databases and files) are extracted by extraction routines, which provide either complete snapshots or differentials of the data sources. And the transformation process in ETL has two main categories: 1) filtering or cleaning operations, like the check

for primary or foreign key violations and 2) transformation operations, for example, transforming the date format to a uniform format based on the data warehouse requirements; Or, for "cost" attributes, transform to a uniform unit of money.

[28] presents a Unified Modeling Language (UML) based approach to accomplish the conceptual modelling of the ETL processes. It presents the design of the ETL process contains six tasks: 1) select the sources for extraction; 2) transform the sources: filtering data, converting codes, performing table lookups, calculating derived values, transforming data formats, etc; 3) joining the data sources; 4) select the target to load; 5) map source attributes to target attributes; 6) load the data. In their approach, the authors defined several ETL mechanisms as: 1)Aggregation: Aggregates data based on some criteria; 2) Conversion: Changes data type and format or derives new data from existing data; 3) Filter: Filters and verifies data; 4) Incorrect: Reroutes incorrect data; 5)Join: Joins two data sources related to each other with some attributes; 6) Loader: Loads data into the target of an ETL process; 7) Log: Logs activity of an ETL mechanism; 8) Merge: Integrates two or more data sources with compatible attributes; 9) Surrogate: Generates unique surrogate keys; 10) Wrapper: Transforms a native data source into a record-based data source.

GENUS [29] is an ETL tool that focuses on dealing with the issue of the variety of data in a big data environment. GENUS can deal with a variety of data types: unstructured data, image data, and video data. This tool extracts the web and then transforms them from the raw data to a suitable one to load into the data warehouse. To achieve the purpose of treating the variety of data types, it intervenes transformation phase and divides it into two parts: 1) data cleansing: the only cleaning task in this tool is to delete the data with missing information; 2) extracting main concepts: aiming to extract concepts and metadata from the dataset and stored them in XML files to be loaded to the data warehouse.

# 5  Data Preparation

In the lifecycle of building a machine learning model, training the model just takes a small part of the machine learning process. Michael Stonebraker and El Kindi Rezig, point out that getting high-quality data accounts for 80% time of the process. At the same time, the other parts just take 20%[234]. To get high-quality data, we consider two factors: 1) data preparation; 2) feature engineering. Data preparation is such an important issue because a machine learning model is acutely sensitive to the quality of data. A low quality data set will have a negative effect on the machine learning model. In a word, this data preparation step plays a crucial part in machine learning, which mainly aims to improve the quality of data in the warehouse.

## 5.1 Data Quality Checking

Data quality of training data has significantly impacted the performance of machine learning tasks, even the complexity of machine learning model. Several types of common data quality problems include inconsistent data conventions amongst sources; data entry errors such as spelling mistakes inconsistent data formats, missing, incomplete, outdated or otherwise incorrect attribute values, data duplication, and irrelevant objects or data.

An extensive data quality survey [32] identifies the key challenges of big data quality evaluation, and also surveys, classify, and discusses the most recent work on big data management. This survey points out that the Data Quality Dimensions (DQD's) usually can be divided into four categories: intrinsic, contextual, representational and accessibility. This survey's classification of data quality in big data extracts major research trends related to Big Data quality, identifies what has been addressed so far in its quality management, and what needs further exploration to reach its full potential. In addition, A survey of Data Quality Dimensions [40] presents the specific definitions along each dimension. The definitions of several mainly dimensions are shown in Table 1.

**Table 1** Caption text

| Dimension | Definition |
|---|---|
| Timeliness | Timeliness is about how long the data was recorded and the information instability |
| Timeliness | Timeliness is about how long the data was recorded and the information instability |
| Currency | Currency is about the degree of a data update |
| Consistency | The extent to which data is presented in the same format and compatible with previous data |
| Accuracy | Accuracy is a measure of the proximity of a data value |
| Completeness | The ability of an information system to represent every meaningful state of the represented real-world systems |
| Duplication | The extent to the mount of unwanted duplication data |
| Security | The extent to which access to information is restricted appropriately to maintain its security |
| Believability | The extent to which information is regarded as trustworthy and credible |
| Objectively | The extent to which information is unbiased, unprejudiced and impartial |
| Relevancy | The extent to which information is applicable and helpful for the task at hand |

## 5.2  Data Cleaning

While data cleaning is a process of improving data quality by removing errors in the dataset, some results show that data cleaning can significantly affect the ML model. Data quality and data cleansing are two closely related topics. Data quality checking targets data cleaning; while data cleaning aims to solve the data quality problems.

A tutorial [18] indicates that data cleaning approaches can be categorised into quantitative cleaning and qualitative cleaning. Quantitative problems are detected by statistical characteristics, such as duplication error, attribute error, relevant error and so on. In contrast qualitative problems are always detected by rules, constraints and patterns, which mainly include integrity constraints (ICs). This tutorial also shows the data cleaning techniques in terms of quantity and quality respectively, as shown in Table 2.

**Table 2**  Comparison of Data Cleaning Techniques

| Technique | Error Type | Ref. |
|---|---|---|
| Continuous data cleaning | One at a time/Holistic/Rules | [35] |
| Trifacta | Attribute Error | [245] |
| parameter-free | Duplication Error | [38] |
| LEAP | Relevance Error | [39] |

[33] is a survey about cleaning dirty data using a machine learning paradigm for big data Analytics. This survey summarized and compared commercialized data quality management tools. Most of the tools concerns organizing data sets, and clean messy data and very few methods uses machine learning. But they didn't give much importance to big data characteristics, which may lead to big challenge while cleaning data. And then it points out several challenges of data cleaning for big data, which include:

1) **Scalability:** As the increasing of size, data cleaning requires scaling data capacities, which is challenging.
2) **Semi-Structured and Unstructured Data:** Big data may be populated with semi-structured data and unstructured data, these two kinds of data remains unfamiliar with quality problem.
3) **Raising Privacy and Security Interests:** While cleaning data, the most common task is to observe and examine a complete set of raw data value that may be restricted in some domains is a significant challenge, like telecommunication, medicine and finance
4) **Computational Complication for Data Streaming:** Huge data collec- tion from various sensors and user devices causes huge processing power of cleansing actions.

5) **Machine Learning Algorithms:** There is still further research going to improve machine learning related algorithms that will be more suit- able with real world conditions which may contain millions and trillions of components for data cleaning.

## 5.3  Data Merging

Data merging is a process of combining two or more data sets into a single data set. In most cases, this process is necessary when there is raw data that is stored in multiple files, worksheets, or data tables, and the task is to analyse all these data in one dataset. There are three main scenarios of merging [2]:

1) **Digital transformation initiative[246]:** When moving disparate files, such as CSV files, SQL databases and other file formats, to a mature data hosting or processing system, data merge is necessary, which can enable automated workflows, enhancing search capability, controlling information access and so on.

2) **Driving business intelligence:** Data merge usually occurs when combin- ing data residing in different applications, and merge them to prepare for further data analysis and processing, and extracting valuable insights for future predictions.

3) **Integrating data after mergers and acquisitions:** Mergers and acquisi- tions include complex moving parts, and one of the most complicated steps is combining data from different companies into one repository and then making processes compatible with newly merged projects, structures, and workflows.

## 5.4  Data Matching

The term data matching is used to indicate the procedure of bringing together information from two or more records that are believed to belong to the same entity. Data matching has two applications: (1) to match data across multiple datasets (linkage) and (2) to match data within a dataset (deduplication). Some open-source and/or freely available data-matching techniques are listed in Table 3.

## 5.5  Exploratory Data Analysis

Exploratory data analysis (EDA) is used by data scientists to analyse and investigate data sets and summarize their main characteristics usually by employing data visualisation methods. It helps determine how best to manip- ulate data sources to get the answers for specific tasks, making it easier for data scientists to discover patterns, spot anomalies, test a hypothesis, or check assumptions.

---

[2]https://dataladder.com/merging-data-multiple-sources/

**Table 3** Comparison of Data Matching Techniques

| Technique | API | GUI |
|---|---|---|
| Dedupe[1] | Python | No |
| fastLink[2] | R | No |
| FEBRL[3] | Python | Yes |
| JedAI[4] | Java | Yes |

[1] https://github.com/dedupe io/dedupe

[2] https://cran.r-project.org/web/packages/fastLink/index.html

[3] https://sourceforge.net/projects/febrl/

[4] https://github.com/scify/JedAIToolkit

Paper [41] presents the exploratory data analysis detail and also states its role in the data analysis domain. It summarised the process of EDA, which contains five main steps:

1) **Identification of Variables:** This step is about sorting the variables and their data type for analysis. Input variables and the Output variables as the target are identified.
2) **Univariate & Bivariate Analysis:** This step helps identify the correla- tion between the attributes.
3) **Detecting Missing Values:** Missing and null data can produce inaccurate results using the model.
4) **Outlier Analysis::** Abnormal distance from other values in the dataset is observed in the dataset.
5) **Visualization:** After cleansing and preparing the data using heatmaps, correlation matrix etc.

# 6 Feature Engineering

Feature engineering is the task of improving predictive modelling performance on dataset by transforming its feature space by selecting, manipulating and transforming raw data into features. Feature engineering can produce several new features that are not in the original dataset, which improves the model's performance.

## 6.1 Feature Selection and Extraction

In the era of ML, handling huge amounts of high-dimensional data has become one of the most challenging tasks that data scientists are facing. Dimension reduction is one of the techniques to pre-process the high-dimensional data for machine learning, which can be categorized into two parts: feature selection and feature extraction. Feature selection directly selects a subset of features using specific methods. On the other hand, feature extraction projects the high-dimensional data into a new feature space, which often has a much lower dimension.

A paper [9] of Computing Surveys, written by Jundong Li, et al. in 2017, provides a comprehensive and structured overview of recent research in feature selection. Compared to other feature selection surveys, this paper reviews the feature selection from a data perspective. This paper categorizes the data types of machine learning into four groups:

1) **conventional data:** For conventional data, feature selection algorithm was grouped as similarity-based, information-theoretical-based, sparse-learning-based, and statistical-based methods. Similarity-based method is a family of methods that assess the ability of features to preserve the similarity of data, while information-theoretical methods are a family that exploits different heuristic filter criteria to measure the importance of features. As most information-theoretic principle can only be applied to discrete variables, information-theoretical methods just can be applied in discrete data. Sparse-learning-based method can reduce specific feature coefficients to nearly zero so that the corresponding features can be removed. At last, statistical-based methods rely on various statistical measures instead of learning algorithms.

2) **structured data:** This type of feature selection methods will consider about the inherent structure of features, such as temporal smoothness, overlap groups, trees and graphs and so on. Combining the knowledge of inherent structure features may help scientists to find more relevant features.

3) **heterogeneous data:** Heterogeneous data is about linked data, multi-sources, and multi-view features. It is challenging to exploit relations among data instances. [13] proposed feature selection algorithm to linked data. for multi-source feature selection, it aims to select features by integrating multiple sources.

4) **streaming data:** There are more streaming data in real-world applications. The size of streaming features is unknown. Thus when faced with stream data, there is an online feature selection algorithm [14] for binary classification, and [15] is a method that selects features when the data are unlabelled. A typical algorithm for streams features selection is to determine whether to accept the most recently arrived feature; if features are added to the selected feature set, then it should figure out whether to discard existing features.

[16] have a briefly highlighted the feature selection and extraction methods. This paper divided feature extraction methods into two parts: unsupervised feature extraction and supervised feature extraction. One of the most popular unsupervised feature extraction methods is Principle Component Analysis (PCA). PCA extracts linear combination features by decomposition of the covariance matrix that after data standardization. The other method to extract features from linear combination features named Independent Component Analysis (ICA)[215], which idea is to extract nonredundant features without

any orthogonality constraint. On the other hand, Non-negative Matrix Factorization (NMF) is a method that decomposes the non-negative features space by matrix factorization. As for supervised feature extraction, it is not as necessary as unsupervised feature extraction because supervised feature extraction is functional overlapping with some machine learning tasks (such as classification). Linear Discriminative analysis (LDA) is the most classic method for supervised feature extraction. LDA tries to extract linear features that can maximise the distances between target labels/classes and minimize the within-label/class data variance.

## 6.2   Feature Construction

Unlike feature selection and feature extraction, which mainly aim at reducing the dimension of feature space, feature construction involves transforming the original features to generate a set of more powerful features that can improve model prediction's performance.

A survey of feature construction methods [96] presents the research in this area in the past 20 years. Most of the discussion in this survey would mainly focus on the use of feature construction for improving prediction performance since it is often the more challenging and less understood aspect. In this survey, it summarised that any feature construction method can be thought of as performing the following.

1) Start with an initial feature space $F_o$ (manual feature construction)
2) Transform $F_o$ to construct a new feature space $F_N$ (feature transformation)
3) Select a subset of features $F_i$ from $F_N$ (feature selection). Then verify the usefulness of $F_i$ for the prediction task based on some utility criteria. If criteria is achieved, then perform step 3 again. If not, set $F_T = F_i$.
4) $F_T$ is the newly constructed feature space.

[96] have summarised that the feature construction methods can be categorised into several types:

1) **Decision Tree Related:** new features are constructed by combining pairs of features in the exisiting feature space using negation and and operators;
2) **Genetic Programming:** An evolutionary algorithm based technique that starts with a population of individuals, evaluates them based on fitness functions and constructs a new population by applying a set of mutation and crossover operators on high scoring individuals and eliminating the low scoring ones;
3) **Annotation Based Approaches:** Annotation-based approaches allow users to provide domain knowledge in the form of annotations along with the training examples. The feature space is then learnt based on these annotations. Thus eliminating the need for defining operators;

## 6.3  Feature Scaling

Feature scaling is a method used to normalize the original features into a specific range of features, such as normalize the data into the range of 0 to 1. Some machine learning model would not work properly without normalization, such as some classification model need to calculate the Euclidean distance of data points. Therefore, after normalizing, each feature will contribute approximately proportionately to the final distance. Another reason for feature scaling is that gradient descent converges much faster with feature scaling than without it.

The mainly two feature scaling techniques are Standardisation and Normalisation. Normalization scales the features to a predefined range (normally the 0–1 range), independently of the statistical distribution they follow. It does this using the minimum and maximum values of each feature in our data set, making it a bit sensitive to outliers. The equation of normalization is: $X_{normalised} = \frac{X - X_{min}}{X_{max} - X_{min}}$

While standardisation takes our data and makes it follow a Normal distribution, usually of mean 0 and standard deviation 1. It is best to use it when we know our data follows an standard distribution or if we know there are many outliers. The standardized equation is: $X_{standarised} = \frac{X - \mu}{\sigma}$ , where $\mu$ is the mean value of the variables, and $\sigma$ is the standard deviation of the variables.

## 6.4  Data Labelling

Data Labelling refers to the process of adding tags or labels to raw data such as images, videos, text, and audio. There are various Labelling approaches. Depending on the problem statement, the time frame of the project and the number of people associated with the work. [235] presents comprehensively the data Labelling techniques, which can be categorized as: 1) Use existing labels: The idea of this category is to learn from the labels to predict the rest of the labels; 2) Crowd-based: a set of techniques that are based on crowdsourcing; 3) Weak labels: it is a kind of labels that are less than perfect labels. But generating weak labels should be done in large quantities to compensate for the lower quality. Table 4 shows the comparison of data Labelling techniques in three categories.

**Table 4**  Comparison of Data Labelling Techniques

| Category | Approach | Techniques |
|---|---|---|
| Use Existing Labels | Self-labelled | [236] |
| | Label propagation | [237] |
| Crowd-based | Semi-supervised+Active  learning | [240] |
| | Crowdsourcing | [242] |
| Weak supervision | Data programming | [243] |
| | Fact extraction | [244] |

## 6.5  Feature Imputation

Feature Imputation imputes missing features on dense instances using user-specified methods or values. Three possible missing feature mechanisms[103]: Missing Completely at Random (MCAR), Missing at Random (MAR), and Not Missing at Random (NMAR). If missingness does not depend on any information in the dataset, then it means that data is missing completely at random, while missing at random means that missingness depends on observed information but not depends on missing values. As for not missing at random, it indicates that missingness is dependent on unobserved data rather than observed data.

[104] shows that there are two types of method that are applied to impute missing features: single imputation and multiple imputation. Firstly, single imputation is about imputing one plausible value for each missing value. Some popular single imputation methods are as follows:

1) **Imputation with the constant:** The missing values are imputed by the constant. In the case of categorical variable, it could replace all missing values with "Missing" or constant values.

2) **Mean Imputation:** This method imputes the missing values with sample mean on the median, or mode depending on the distribution or distribution of the data. This method is easy to implement and understand but it has its own drawbacks. It replaces the missing value with sample mean or median or mode depending or distribution of the data. Study[148] shows that the limitations of mean imputation are almost absent if less than 10% of the data is missing and when the correlations between the variables are low. Two techniques similar to mean imputation are median and modus imputation. These methods were invented to account for imputation of not normally distributed data, but they suffer from the same limitations as mean imputation and are therefore not very popular methods of data imputation.

3) **Imputation with distributions:** In this approach, missing values are replaced by random values from the known distribution. The imputed value does not change the shape of the distribution.

4) **Regression Imputation:** In this method, the missing value is replaced by predicted data using regression based on non-missing data of other variables. This method is based on the assumption of linear relationship between the attributes. But most of the time relationship is not linear, so replacing the missing value using regression will bias the model.

5) **K-nearest neighbors (KNN) Imputation:** In this method, missing val- ues are imputed by copying values from similar records in the same dataset. The similarity of the two attributes is determined using a distance function. The creation of predictive model for each attribute is not required, but it also has disadvantages. It is very time-consuming in analyzing a large dataset. Choice of k value is also critical. Several variants and packages of this method have been proposed in recent years[149, 151].

The idea of multiple imputation[159] is to create a number of data copies, each of which has suitably imputation for the missing values , and each complete dataset is analysed independently. This method follows three steps: (1)Imputation: missing values are imputed by generating one missing data "m" times rather than just once, which led to 'm' different complete datasets after imputation. (2) Analysis of each dataset: each of the 'm' datasets generated in step (1) is analysed. (3) Pooling: Results obtained from each analysed datasets are consolidated.

Multivariate Imputation by Chained Equations (MICE) is a popular method for multiple imputation. MICE package [97] is an R based package that imputes incomplete multivariate data by chained equations.

Besides the classical methods mentioned above, there are several novel feature imputation methods that have been proposed in recent year. GAIN [185] is an imputation algorithm that adapts the Generative Adversarial Nets (GANs). In GAIN, the generator's goal is to accurately impute missing data, while the discriminator's goal is to distinguish between observed and imputed components. SICE [187] is a new technique that is a hybrid approach of single and multiple imputation techniques. It has proposed an extension of MICE algorithm in two variations to impute categorical and numeric data.

# 7 Model Training

This section summarises the characteristics of model training. The initial of this part illustrates four major groups of ML models before introducing how to select suitable methods from given problems. We address the model selection from both academic methods and enterprise solutions. In the end, we summarise the parameter tuning methodologies, and the major challenges relating to model training.

## 7.1 Algorithm selection

Researchers and engineers created many algorithms over years. Some algorithms are suited for tableau data, others fit for sequence data such as text or music, and some algorithms target numeral data or image data. In summary, there are divided into four different categories: Supervised learning model, Unsupervised learning model, Semi-supervised learning model, and Reinforcement learning model.

1) **Supervised learning model:** In the supervised learning model, the historical data are labelled as the ground truth and an algorithm learns to predict the output from the input data. The learning model adjusts its parameters until the model has been fitted well. According to the label types, there are two basic categories of problems this type of model solved: classification and regression. The classification problem is an algorithm that accurately organizes test data into specific categories for its most effective use. Common classification algorithms are Logistic regression, Naïve

Bayes, Support vector machine (SVM), K-nearest neighbor (KNN), random decision forest. Fitting a regression model is the process of identifying the relationship between dependent variables and independent variables on the performance of future values. Besides, classification is used to predict both numerical and categorical data with four types: binary classification, multiclass classification, multilabel classification, and imbalanced classification, whereas regression is used to predict numerical variables only. Table 5 for a comparison of different algorithms.

**Table 5** Comparison of Supervised Learning Algorithms

| Name | Training Time (quick) | Data Size (Big data) | Explainable (Easy) | Accuracy (High) | Linearity |
|------|------------------------|----------------------|--------------------|-----------------|-----------|
| Linear Regression | Yes | No | Yes | No | Yes |
| Logistic Regression | Yes | No | Yes | No | Yes |
| Naive Bayes | Yes | Tes | No | No | No |
| KNN | Yes | Yes | Yes | No | No |
| Decision Tree | Yes | No | Yes | No | No |
| SVM | Yes | No | No | Yes | Yes |
| Random Forest | No | Yes | No | No | No |
| Boosting Tree | No | Yes | No | Yes | No |
| Artificial Neural Network | No | Yes | No | Yes | No |

2) **Unsupervised learning model:** Compare with the supervised learning model, all input data are unlabelled in the unsupervised learning model. The unsupervised learning algorithms learn the inherent structure from data that help solve clustering and association rule problems. In some scenarios, labelled datasets are extremely hard to come by and it require intensively human review. Thus, with a lot of unlabelled data, the learning algorithm attempts to make an appropriate prediction from the unlabelled dataset.

There are four types in unsupervised learning models. (1) The most common type within unsupervised learning models deal with clustering problems. One of the most popular method is the K-means algorithm. (2) The principal component analysis (PCA) has been popularised method to convert the high dimensional data into lower dimension space and it is also deep connection with the k-means algorithm. (3) Next, association rule leaning model is to find group of features that are commonly associated with. The Apriori algorithms [170] are a very efficient method for the market basket analyses. Apriori algorithms explore frequent itemset mining on the concepts of association rules. (4) the Autoencoders [171] is a neural network that intends to do two things. Firstly, it compresses its input data into next layer with a lower dimension. Next, it uses this data representation to reconstruct the original input. Table 6 describes of main differences between popular unsupervised learning algorithms.

**Table 6**  Comparison of Unsupervised Learning Algorithms

| Name | Type | Training Time (quick) | Data Size (Small) | Explainable (Easy) |
|---|---|---|---|---|
| K-means | Clustering | Yes | Yes | Yes |
| K-modes | Clustering | Yes | Yes | Yes |
| Gaussian Mixture | Clustering | Yes | Yes | No |
| DBSCAN | Clustering | Yes | No | No |
| Mean-Shift Clustering | Clustering | Yes | No | No |
| Hierarchical Clustering | Clustering | No | Yes | Yes |
| PCA | Dimension Reduction | No | No | Yes |
| SVD | Dimension Reduction | No | Yes | No |
| LDA | Dimension Reduction | No | Yes | Yes |

3) **Semi-supervised learning model:** Semi-supervised learning models operate in the situation of some data is labelled but most of it is unlabelled. This type of algorithms uses the classification process to identify data label and the clustering process to categorise test data into distinct parts [198]. One of using semi-supervised learning models is medical image labelling because it is especially difficult to label medical images because it is not only time-intensive for a large dataset, but the data collection processes protecting by law or regulations. A situation for this type of machine learning are to label million or billion pictures. Few of them marked by users, but most labels are unknown. The algorithms train the small set of labelled data to generate a best-trained model. Then this model labels those unlabelled data. After that, the results of labelling are considered to be pseudo-labelled data. The pseudo-labelled data combined with labelled data to create a mechanism with both supervised and unsupervised learning characteristics.

The popular semi-supervised learning algorithms are self-training [181], Semi-supervised support vector machines [182] mixture models [118] and Semi-supervised generative adversarial networks(SGAN)[184].

4) **Reinforcement learning model:** In brief, reinforcement learning [188] is considered as from a state S, taking action A in an environment results to maximise some notion of reward. This mechanism helps participants to improve their performance for the next iteration. Next, it provides a reward signal, but the reward feedback is delayed. In the learning process, there is no supervisor to guide the action is right or wrong. The process is sequential, and time really matters for reinforcement learning.

Dynamic programming, Monte Carlo Methods and Temporal Difference are the three sub-components of reinforcement learning. Sutton and Barto provide a comprehensive introduction in [5]. The emergence of deep reinforcement learning that combines artificial neural networks with the reinforcement learning algorithms has been adopted in wide range of areas [193].

Microsoft and scikit-learn provided common machine learning algorithm maps to help choose the right algorithms for a given data and problem that they try to solve.

## 7.2   Model selection

There is not exist a universal model is suitable for any data and prediction problems. Thus, we need the model selection procedure to determine which model is the best model for a group of candidates. This section reviews different methods that can be used for model selection.

### 7.2.1   Resampling Methods

Resampling methods are a set of methods to repeatedly select samples from given training sets and to choose the final model. Three of the most popular resampling methods are cross-validation, bootstrapping and Random Split. A summary of these methods is presented in Table 7.

1) **Cross-Validation (CV):** In general, CV is the most popular method that works for both model selection and model evaluation because it results in less biased than other methods. This method was studied in a series of papers for many kinds of machine learning problems [57, 61–63].

2) **Bootstrap:** The bootstrap method is a process of drawing a repeated sample from a given dataset [195]. One important point is a bootstrap sample can be selected more than once because any selected sample needs to be sent back into the given dataset.

3) **Split:** This method splits the population to train the dataset, validation dataset and test dataset. It is a common method to minimise overfitting and underfitting that is widely used in training machine learning models [196]. Two of the split methods are Random split and Time-Based Split [197].

**Table 7** Overview of Resampling Methods

| Type | Description | Advantage | Disadvantage |
|------|-------------|-----------|--------------|
| HoldOut CV | Split the dataset into 2 (training and testing) parts or 3 (training, validation and testing) parts. | 1. Run once only 2. Fast for small dataset | 1. Bias for imbalance dataset 2. Hard to train a model with a large dataset |
| K-Fold CV | Split dataset into K partitions. K-1 part is used for training and 1 part as the validation set. Then iterate over it K times | Guarantee each sample is used in both training set and validation sets. | 1. K is hard to decide 2. Not works for the time series dataset 3. Bias for imbalance dataset |
| Stratified K-Fold CV | Split dataset into K parts and each part can approximately representative of the whole dataset | Fit well for imbalanced data | 1. Supervised learning only 2. Not works for the time series dataset |
| Leave P Out CV | Using P samples as validation and remaining samples as the training set. Then iterate over the above steps until the whole dataset finishes both roles. | Guarantee each sample is used in both training set and validation set | 1. Intensive computation 2. Not works for the time series dataset 3. Bias for imbalance dataset |
| Monte Carlo CV | The whole dataset gets randomly partitioned by training set and validation set with uniform distribution | More flexible in data partition | 1. the same partition to be chosen more than once 2. Not works for the time series dataset 3. Bias for imbalance dataset |
| time-series CV | Split dataset on a forward rolling basis | works well for dependent dataset | Bounded by the order of data |

### 7.2.2 Probabilistic measures

The probabilistic measures are are applied to parametric based machine learning models only and they are considered model performance with model complexity.

1) **Akaike Information Criterion (AIC):** Akaike [67] was the first to propose a general criterion for selecting models estimated by maximum like- lihood. AIC deals with the trade-off between model prediction loss and a correction term that measures the complexity of the model.

2) **Bayesian Information Criterion (BIC):** BIC [68] is one of the widely known model selection principles. It selected the model with lowest BIC. The Major difference from AIC is BIC replaced the Constance value in the penalty part. Thus, this modification gave more penalization to the model complexity.

Kuha [66] examined and compared AIC and BIC on the performance of the model selection. Except AIC and BIC, there are some other widely used approaches for model selection. Such as, Minimum Description Length (MDL) [69], Structural Risk Minimization (SRM) [70], Hannan and Quinn criterion(HQ) [71], Bridge criterion(BC) [72]. All introduced approaches typically work efficiently on some subset of the model selection problems.

## 7.3  Hyper-parameter tuning

We explore hyper-parameter optimization in this section. Hyper-parameter tuning is about choosing meta-value of a model to get a more accurate prediction for a ML algorithm. The tuning step adjusts the model interactively. Each of these iterations is so call a training step. Tuning the Hyper-parameter is more art than science [44]. Bergstra and Yu [85] reviewed popular algorithms for hyper-parameter tuning. Here list a few methods:

1) **Grid search:** Grid search is the most straightforward search algorithm that searches every given combination of hyper-parameters and evaluates each model. Grid search leads to good accurate results but it cost a lot of computational resources [74].
2) **Random search:** Random search [75] keeps similar accuracy as grid search, but it is more efficient than the grid search method for optimizing a machine learning model.
3) **Bayesian Optimization (BO):** Tuning the hyper-parameter with Bayesian optimization [76] is an approach that uses the Bayes theorem to update the probabilities of a hypothesis in order to find the best parameters from a search space.
4) **Evolutionary Optimization:** Evolutionary hyper-parameter tuning fol- lows a process inspired by the concept of evolution [7]. Lorenzo [73] proposed a way to find parameters with the particle swarm optimization (PSO). Bouktif [77] applied the genetic algorithm (GA) to optimize a deep learning model.

# 8  Model Evaluation

There is no doubt that the previous steps mainly focus on the training and changing of the different models based on the performance and results of training data. Extra efforts and contributions are required before the publication. According to the research by IBM [106], currently most of the models are designed by academic teams, which is not suitable for different kinds of industrial uses. According to [136], Quality Assurance(QA) for ML system is much more difficult than other kinds of software products because of the special feature of ML, i.e. low explainability, low controllability and low predictability. Different testing and operations are needed to investigate the performance of the system under different usage and running circumstances. By combining the A-ARAI principles of ML model, which are allowability, achievability, robustness and improvability, we established some sub-chapters for the evaluation.

## 8.1   Model testing

Model testing is one of the most important parts of this section. This is the metric to evaluate the achievability of the ML model, the performance of the model and how the model manages target tasks and satisfies clients. Usually, this involves tests among input datasets processing, functional implementation, output figure relevance, output result coverage ratio and other test elements. Generally, the testing part should be involved with different kinds of input values; this requires some real and latest datasets from different stakeholders. There are several sub-characteristics for the testing.

1. Model system testing: Despite the traditional testing methods for system evaluation like unit tests, regression tests and integration tests. The system testing for ML model gains more attention to the interaction and appearance of the client and stakeholders. The achievability, robustness and avoidability of the ML model will be tested among the connections and feedback among hardware, management system and human.

2. Model software testing: The analysis team from Microsoft [48] recommends automated testing pipeline and an integrations of the model with some human supervisors for software engineering testing. We divided the testing into three main parts.

1) **Reliability:** The reliability includes robustness, error tolerance and recoverability [129]. To analyse the capability of handling failure and error cases, test elements are recommended i.e. hazard detection, failure avoidance and average recovery or restart time.

2) **Performance and efficiency:** The efficiency of the ML model is judged by both time behavior and resource utilisation to evaluate the system performance under different hardware environments.

3) **Maintainability:** The maintainability of the ML model consists of four parts: normativity, convergence, changeability and verifiability. These metrics [132] will define the quality of the model by inspecting the code to figure out whether the model is easy to change and develop in the future.

3. Model data testing: Data testing for ML models mainly focuses on the data wrangling, variation, multicollinearity and other changes of the datasets [137]. Tests are operated among different kinds of input data for the model, including error datasets, transformed datasets, auto-generated datasets, and datasets from different original resources. This level of tests is organized to evaluate the model performance based on different input figure. Principal Component Analysis(PCA) is generally adapted to investigate the possibility of data storage requirement deduction for the upcoming deployment stage.

4. Model component testing: The tests of each model component are carried out

with other different ML components. Usually, we need to compare the result and statistics of the selected ML component with other ML[105] or non-ML components. During this stage, testers and developers could find out whether each of the model components is suitable for the goal or not.

## 8.2  Result evaluation

The result of the model is defined as the combination of different statistics collected from the model as well as outputs of particular algorithms, which could be used to analyse the model. Although there is no widely used algorithm to evaluate all kinds of models, we tried to gather and combine all the different evaluation metrics from [129, 131, 133, 134]. Initially, we summarised the most basic and widely used methods, displayed as Table. 8

For some particular cases, [138] has proposed three upgraded metrics used for ML evaluations:

1) **Balanced F-Score**: This is a kind of upgraded version of the F-score, which could reflect the accuracy of testing cases. On the other hand, here we try to give equal weights for the harmonic mean of both tests precision and recall values to reach the balance of different data sources.
2) **Hamming Loss**: Hamming loss is a valuable evaluation method for multi-label classification models. This shows the fraction of incorrect predictions, which could be obtained as the ratio of error numbers.
3) **Silhouette Coefficient**: The SC is a measurement for finding the best model among K-mean clusters. This equation of the score is below where a is the average of the distance among clusters and b is the average sample distances from all points assigned to the closest cluster.
4) **Chi Squared Feature Selection Tests (CSFST)**: The CSFST is a hypothesis test used for categorical variables, which analysis the null hypothesis that is independent of each other. In other words, the Chi test could determine the dependence of some selected features on the classifier. The following is the calculation of the equation in which larger than x indi- cates less probability and the model does not have a good representation of the population
5) **Adjusted Rand Index**: The ARI is the adjusted version of the rand index, which uses the expected rand index value to get the correct score. This could be used to compare the predictions and actual results.
6) **Area Under Receiver Operating Characteristic Curve (AUC)**: AUC is an upgrade of ROC, which is used to measure the performance of classifiers' distinguishment. Generally, this curve is produced by the parameters of actual labels as well as the predicted scores from the classifier.

From [134], the author also uses the Median Absolute Percentage Error (MedAPE), which takes the median value over the mean absolute percentage error values after cross-validation. This method could display the fluctuation of the results, which lower value presents higher model accuracy.

**Table 8** Comparison of Evaluation Metrics

| Evaluation Metrics | Description |
| --- | --- |
| Accuracy | Accuracy is the most generally used method that analysis the model with the correct percentage of testing data. |
| Mean Squared Error | A popular method used to analysing the performance of the linear model, using the squared roots of mean errors. |
| R2 Score | R2 is used to investigate a linear model especially when the target statistics includes different equivalent classes. |
| Confusion Matrix | A common way to plot four different blocks including the true positive (tp), true negative (tn), false positive (fp) and false negative (fn) predictions. |
| N-Fold cross | A kind of cross-validation between different groups of training dataset and testing dataset when lacking enough figures to train and test. |
| Analysis of Variance | This is a method from statistics, which could illustrate the overall fluctuations of the results, also used for linear regression models. |
| Decision distance matrix | A method for dividing different decision groups of the result is used for Learning Trees (LT). |
| P-value | P-value is from statistics to evaluate the relationship between the predictions and the model. |
| F beta score | F beta is used as an F1 score which could consider both the accuracy and recall of the classification models. |
| Kolmogorov-Smirnov statistic | Ks statistic is used to plot and illustrate the separation between prediction distributions for positive and negative classes. |
| Receiver Operating Characteristics (ROC) curve and AUC score | ROC is a graph using a true positive rate (TPR) and false-positive rate (FPR) for performance comparison, which commonly used in Medical Sciences cases. AUC is the value extracted from the plot to rank the correlation between predictions and targets. |

In addition, [139] gives some specific methods to evaluate the performance safety of the ML model for particular cases. The performance does not only depend on the accuracy but also on wrong positive values.

## 8.3 Business value review

To reach the goal of a thoroughly designed and developed ML program, the developer may need to consider several systems related future business plans and strategies. Despite some particular academic projects, most of the models require previous business value reviews before deployment. Here we mainly summarised some possible aspects of the role of developers and programmers of the model and system.

**Business value:** In the book Prediction Machines, professor Ajay has mentioned that both business leaders and developers should understand their project's business value, including the power and limits of the predictions provided by the model and the inside algorithms before deployments. The author [142] has suggested several possible business values like the advantage of ML predictions, including reducing costs and promoting extra orders. In this case, we have summarised several questions:

1) Is the model operation cheap enough to attract commercial investment?
2) Will the cheap predictions of the ML system change the way organisations work?
3) Can the recommendation and suggestion bring extended benefits like extra purchases and orders?

**Potential expense:** Some potential expenses costs should be considered during the business review processing, like data value. In academic aspects, scientists most focus on the performance of different models among well-spread datasets such as ImageNet, CLEVR, or SQUAD. However, industrial programs require more realistic testing and implementation among real-world data. Hence, a great job is needed to obtain the correct data for training and maintaining machine learning algorithms. The authors of [124] have underlined that gathering quality data is costly and time-consuming, which should be included as a potential expense. Companies may involve a trade-off between the cost of acquiring essential data and cheaper predictions that ML systems could provide.

**Local regulation:** Apart from traditional software systems, many new policies and regulations are published for AI systems. In this case, developers should also be concerned about the relative local regulations for the system. There are some new rules like some extra taxes and new individual privacy regulations, particularly for AI systems. Hence, designers should combine these kinds of principles with the business approach of the target system. To figure out whether the current model and the algorithm data are suitable or not, the following questions should be considered among regulation aspects:

1) Do we have enough rights to use the data for training and testing?
2) Is there any extra costs or taxes for the algorithms and models used in the ML system?
3) Does the model provide legal outputs and protect individual privacy successfully?

**Future development & updates & maintenance:** Like traditional software systems, ML programs also need to have further updated versions and maintenance. This topic is also involved in the business view as the update of database and maintenance of the system. The IDC report provided several manufacturing cases and informed that these two elements are the largest expenses in the further development of ML models. Hence, developers should be aware of the plans of the system to avoid any business obligations.

## 9 Model Evaluation

There is no doubt that the previous steps mainly focus on the training and changing of the different models based on the performance and results of training data. Extra efforts and contributions are required before the publication. According to the research by IBM [106], currently most of the models are designed by academic teams, which is not suitable for different kinds of industrial uses. According to [136], Quality Assurance(QA) for ML system is much more difficult than other kinds of software products because of the special feature of ML, i.e. low explainability, low controllability and low predictability. Different testing and operations are needed to investigate the performance of the system under different usage and running circumstances. By combining the A-ARAI principles of ML model, which are allowability, achievability, robustness and improvability, we established some sub-chapters for the evaluation.

### 9.1 Model testing

Model testing is one of the most important parts of this section. This is the metric to evaluate the achievability of the ML model, the performance of the model and how the model manages target tasks and satisfies clients. Usually, this involves tests among input datasets processing, functional implementation, output figure relevance, output result coverage ratio and other test elements. Generally, the testing part should be involved with different kinds of input values; this requires some real and latest datasets from different stakeholders. There are several sub-characteristics for the testing.

1. Model system testing: Despite the traditional testing methods for system evaluation like unit tests, regression tests and integration tests. The system testing for ML model gains more attention to the interaction and appearance of the client and stakeholders. The achievability, robustness and avoidability of the ML model will be tested among the connections and feedback among hardware, management system and human.

2. Model software testing: The analysis team from Microsoft [48] recommends automated testing pipeline and an integrations of the model with some human supervisors for software engineering testing. We divided the testing into three main parts.

1) **Reliability:** The reliability includes robustness, error tolerance and recoverability [129]. To analyse the capability of handling failure and error cases, test elements are recommended i.e. hazard detection, failure avoidance and average recovery or restart time.

2) **Performance and efficiency:** The efficiency of the ML model is judged by both time behavior and resource utilisation to evaluate the system performance under different hardware environments.

3) **Maintainability:** The maintainability of the ML model consists of four parts: normativity, convergence, changeability and verifiability. These metrics [132] will define the quality of the model by inspecting the code to figure out whether the model is easy to change and develop in the future.

3.Model data testing: Data testing for ML models mainly focuses on the data wrangling, variation, multicollinearity and other changes of the datasets [137]. Tests are operated among different kinds of input data for the model, including error datasets, transformed datasets, auto-generated datasets, and datasets from different original resources. This level of tests is organized to evaluate the model performance based on different input figure. Principal Component Analysis(PCA) is generally adapted to investigate the possibility of data storage requirement deduction for the upcoming deployment stage. 4.Model component testing: The tests of each model component are carried out with other different ML components. Usually, we need to compare the result and statistics of the selected ML component with other ML[105] or non-ML components. During this stage, testers and developers could find out whether each of the model components is suitable for the goal or not.

## 9.2  Result evaluation

The result of the model is defined as the combination of different statistics collected from the model as well as outputs of particular algorithms, which could be used to analyse the model. Although there is no widely used algorithm to evaluate all kinds of models, we tried to gather and combine all the different evaluation metrics from [129, 131, 133, 134]. Initially, we summarised the most basic and widely used methods, displayed as follows:

For some particular cases, [138] has proposed three upgraded metrics used for ML evaluations:

1) **Balanced F-Score**: This is a kind of upgraded version of the F-score, which could reflect the accuracy of testing cases. On the other hand, here we try to give equal weights for the harmonic mean of both tests precision and recall values to reach the balance of different data sources.

2) **Hamming Loss**:Hamming loss is a valuable evaluation method for multi-label classification models. This shows the fraction of incorrect predictions, which could be obtained as the ratio of error numbers.

3) **Silhouette Coefficient**: The SC is a measurement for finding the best model among K-mean clusters. This equation of the score is below where a is the average of the distance among clusters and b is the average sample distances from all points assigned to the closest cluster.

4) **Chi Squared Feature Selection Tests (CSFST)**: The CSFST is a hypothesis test used for categorical variables, which analysis the null hypothesis that is independent of each other. In other words, the Chi test could determine the dependence of some selected features on the classifier. The following is the calculation of the equation in which larger than x indicates less probability and the model does not have a good representation of the population

5) **Adjusted Rand Index**: The ARI is the adjusted version of the rand index, which uses the expected rand index value to get the correct score. This could be used to compare the predictions and actual results.

6) **Area Under Receiver Operating Characteristic Curve (AUC)**: AUC is an upgrade of ROC, which is used to measure the performance of classifiers' distinguishment. Generally, this curve is produced by the parameters of actual labels as well as the predicted scores from the classifier.

From [134], the author also uses the Median Absolute Percentage Error (MedAPE), which takes the median value over the mean absolute percentage error values after cross-validation. This method could display the fluctuation of the results, which lower value presents higher model accuracy.

In addition, [139] gives some specific methods to evaluate the performance safety of the ML model for particular cases. The performance does not only depend on the accuracy but also on wrong positive values.

## 9.3 Business value review

To reach the goal of a thoroughly designed and developed ML program, the developer may need to consider several systems related future business plans and strategies. Despite some particular academic projects, most of the models require previous business value reviews before deployment. Here we mainly summarised some possible aspects of the role of developers and programmers of the model and system.

**Business value:** In the book Prediction Machines, professor Ajay has mentioned that both business leaders and developers should understand their project's business value, including the power and limits of the predictions provided by the model and the inside algorithms before deployments. The author [142] has suggested several possible business values like the advantage of ML predictions, including reducing costs and promoting extra orders. In this case, we have summarised several questions:

1) Is the model operation cheap enough to attract commercial investment?
2) Will the cheap predictions of the ML system change the way organisations work?
3) Can the recommendation and suggestion bring extended benefits like extra purchases and orders?

**Potential expense:** Some potential expenses costs should be considered during the business review processing, like data value. In academic aspects, scientists most focus on the performance of different models among well-spread datasets such as ImageNet, CLEVR, or SQUAD. However, industrial programs require more realistic testing and implementation among real-world data. Hence, a great job is needed to obtain the correct data for training and maintaining machine learning algorithms. The authors of [124] have underlined that gathering quality data is costly and time-consuming, which should be included as a potential expense. Companies may involve a trade-off between the cost of acquiring essential data and cheaper predictions that ML systems could provide.

**Local regulation:** Apart from traditional software systems, many new policies and regulations are published for AI systems. In this case, developers should also be concerned about the relative local regulations for the system. There are some new rules like some extra taxes and new individual privacy regulations, particularly for AI systems. Hence, designers should combine these kinds of principles with the business approach of the target system. To figure out whether the current model and the algorithm data are suitable or not, the following questions should be considered among regulation aspects:

1) Do we have enough rights to use the data for training and testing?
2) Is there any extra costs or taxes for the algorithms and models used in the ML system?
3) Does the model provide legal outputs and protect individual privacy successfully?

**Future development & updates & maintenance:** Like traditional software systems, ML programs also need to have further updated versions and maintenance. This topic is also involved in the business view as the update of database and maintenance of the system. The IDC report provided several manufacturing cases and informed that these two elements are the largest expenses in the further development of ML models. Hence, developers should be aware of the plans of the system to avoid any business obligations.

# 10  System Deployment

After the model evaluation phase is model deployment. This step rolls out the selected model into a production environment. Models deployment is a critical task in production machine learning systems. A wrong decisions can destroy a project or leaded to huge maintenance costs. This section reviews the knowledge and skills that are required when deploying ML applications. As we know, machine learning software development is not only about the machine learning model, Baier et al.[95] state the deploying challenges from pre-deployment, deployment and non-technical aspects. Apart from academics, both IBM and Software have suggested that the system's deployment is an extra but essential process among industrial projects. Instead of outstanding accuracy and innovative algorithms, most companies only focus on the actual value that machine learning models could create for their customers. Hence, deploying an ML model to a product is usually considered with a different work than all the previous software works Engineers are seeking ways to secure and balance the deployment between performance and expenses. The whole deployment procedure has been summarised into several steps.

## 10.1 Deployment planning

Initially, developers should have a completed and detailed schedule for the system deployment. Different purposes require different ways of deployment. Thuwarakesh from Stax.inc [119] has already summarised three main types of ML model operating platforms.

1) Deploy for a web application: Running through the internet is the simplest and easiest way for ML systems. Usually, the model runs on a cluster or cloud server and connects with local devices through API. This is an on-demand prediction service that only gives results and suggestions when the user sends requests and data through the web. Developers can develop and train the model local and upload the system using a web framework. The web deployment is cheaper and provides real-time predictions. However, it relays heavy dependencies on external data and cloud services and is hard to conduct online debugs.

2) Deploy batch predictions: Batch predictions are for offline ML systems. This kind of model targets complex situations and jobs with high volume instances. The models are usually more complex than web systems, with more layers and various algorithms. Predictions are called by giving a set of input data. The system's performance depends on local processing power like CPU and storage, which reduces scaling and management issues. In addition, changing or maintaining coding work becomes more efficient.

3) Deploy as embedded models: Embedded model plans are used for customized cases and single target devices, which is the best solution for mobile and IoT devices This kind of model has a more secured environment and is easy to deploy without any system or device compatible issues. Nevertheless, Thuwaraksh suggests that deploying for edge devices may reduce latency and data bandwidth consumption between the communication of clients. However, the edge devices need to have enough power and storage to guarantee performance and accuracy.

## 10.2 Resource preparation

Despite deployment planning, the preparation includes of collecting and configuring of data resources. The quantity and quality of the data resource are essentially important for the commercial operations of the system [120]. Developers should make sure that all the datasets for the system are under privacy regulations [115]. Nevertheless, suitable resources should be approved and confirmed to avoid any discrimination operations from ML systems. Data is one of the most important resources in ML training. Therefore, different types and databases are selected for various kinds of products, like domestic or international; individual or industrial; single or multi-sex. Final checks are required before deploying to the server for investigation of the design and training of the system with the potential targets. This can avoid some severe issues apart from academic malfunction like fairness in AI programs.

## 10.3 System Integration

After containerising the system, integration testing is required for target cases and scenarios to investigate the drawbacks and flaws during development [116]. Integration tests are the last validating procedure before serving. During this stage, the tests are combined for the whole system through the server or deployment platform. Unlike traditional software, the purpose of tests for ML models is to ensure that the system can provide a consistent result and accuracy under any circumstances.

Continuous integration (CI) testing [99] could present the system's performance by presenting the realistic effects of the system instead of digital results and mathematical comparisons between different ML models. Nevertheless, integration tests for ML models are usually harder than normal software testing since it usually takes a longer time to run the entire pipeline end-to-end. Therefore, Google Developer [117] has suggested adopting subset data on training during integration testing. Nevertheless, a test framework could help to achieve the goal of continuous integration testing, which can be carried out easily and efficiently when fixing bugs and capturing new features.

We summarise with several details of integration testing for ML systems. First, The integration testing is between the input figure and output results of the whole system. Second, developers should treat the system from the perspective of clients during integration testing of the deployment stage, like getting the results from API. Last, continuous integration tests are required for updates or changes to the code or model inside the system.

## 10.4 System release

The last section of ML system releasse is launching the system to end-users. By entering this stage, the system has been proved to solve target problems. Deploying a machine learning system not only deals with the source code but also the mathematical models and complex data. Thus, the deployment time was dramatically high compared with traditional applications and it elicits various challenges during the processes.

The system would be provided access to different kinds of users and stakeholders from this step. Usually, beta versions are launched initially, which is required by companies like Amazon and Apple for public testing to collect enough confidence and data before publishing or overtaking the previous systems [154]. Beta testing on customer validation is not an option but an inevitable step. It is hard but can receive enough feedback for the feasibility, viability and desirability of the target system [144]. Overall, launching the system indicates that the system is well-developed, and users could reach them and operate them in different ways. Developers should ensure that users can easily access the system after the launch. From the previous sections, we have introduced several ML system platforms including individual device installations and cloud servers. Most of the public usage of ML systems depends on the operation through cloud servers. For large systems with supportive funding from target companies, developers could consider some large but expensive platforms like Google Cloud, Amazon AWS and Microsoft Azure. On the other hand, for some individuals or small group developers who just aim to test or share their ML systems, we have also gathered several alternative platforms that are free and easy to deploy, like Algorithmia, PythonAnyway and Heroku [3].

To release a product smoothly, Rudrabhatla[98] and Carroll el al. [99] summarised different deployment strategies across organisations. We compared two of the most widely used methods in Table 9.The suitable deployment processes purely depend on the use case and the current situation of the product.

1) **Canary deployment**: Canary deployment is to deploy the new machine learning applications to a small fraction of the users initially, then test the system stability, performance and analysis feedback before rolling it out gradually to the rest of the servers. In this development, there is no switching the whole workload to the new environment because this new application may not be stable or fit with new data due to data drift. Small traffic hit the newly updated application into the production environment, then roll out the rest traffic with multiple steps. Once this deployment reaches a 100 per cent workload, the old environment will be terminated.

2) **Blue Green deployment**: Blue green deployment is another popular deployment method. In this method, a machine learning system transfers user traffic from a Blue version refers to the existing product to a Green version refers to the new release, both of which are running in a production environment. Once all workloads are fully deployed from the blue environment to the green environment, the old version of production can standby in case of rollback or pull from production. After we satisfied the functional and non-functional tests on the green environment, the blue one can be destroyed or switched as the new green environment.

## 11 Model Monitoring

Like other software systems, ML systems also require maintenance for stable performance. Despite skills in the development of ML models and the investigations on ML theories and algorithms, the research from Microsoft [53] has pointed out that most experts and developers lack the knowledge and experience in ML model monitoring including system maintenance and further updates. Here we have summarised several aspects and groups for developers to monitor and manage the ML models.

---

[3]https://www.freecodecamp.org/news/deploy-your-machine-learning-models-for-free/

**Table 9** Comparison of Deployment methods

|  | Canary Deployment | Blue/Green Deployment |
|---|---|---|
| Livability after deployment | Terminated after deployed | Switch from Blue to Green after deployed |
| Workload | small percentage (e.g. 5%) | Fully released |
| Release processes | Gradually increase | Fully released |
| Advantage | Spot issue early Less risky Zero downtime less of computation and storage | Fast and simple in the deployment Easy rollback to the other version |
| Disadvantage | Complex to maintain both versions Make the release slow | Twice of computation and storage that is expensive and complex More work on quality assurance and testing Big impact if anything wrong |

## 11.1    Drift Detection

Apart from traditional programs, where the performance only depends on hardware features like storage and CPU, and ML systems are unique groups of software systems. The accuracy or performance of the system might fluctuate due to the changes in input and 'ground truth'. This kind of change is considered data drift, which leads to a significant reduction of the model outputs. ML drift can be categorized with the data distribution into several different groups including concept drift, prediction drift, label drift and feature drift. Although there are plenty of models evaluating metrics as we introduced in previous sections, these metrics are operated under the concept of 'ground truth' and labels for real-time predictions. Therefore, in case of the training data became outdated, developers should bring with new detections to monitor drifts in ML systems. There are a variety kinds of detectors experts could use to monitor drifts, as shown in Table 10

[118] introduced a kind of Adaptive Window (ADWIN2) that can use to detect distribution change among data sequences. They provide an algorithm to calculate the stable value for each sliding window. ADWIN2 is suitable to monitor time-serious data and concept drifts can be detected as large fluctuations of the window value.

## 11.2    QoM Monitoring

The purpose of monitoring is to make sure that the model works in our predicted way and does its best to evaluate the predictable failures. It is critical to the continued development of machine learning and mainly works to train loss and validate loss trends. The training loss makes the model underfitting and unsuitable for the specific task. In summary, the expectation is that training loss decreases with the time of the training process. The same is true for validation loss. If it increases during the deep training process, the model is overfitting, and the researcher should not continue to work on it. The following reasons make the researcher focused on model monitoring.

**Table 10** Different drift detections

| Detection Metrics | Equation (For two samples) | Description |
|---|---|---|
| Kullback-Leibler (KL) | $D_{KL}(PQ) = \int_{-\infty}^{\infty} p(x) log(\frac{p(x)}{q(x)})dx$ | KL divergence uses the difference of data distribution, which is suitable for time-serious, image and text datasets. |
| Kolmogorov-Smirnov(KS) | $D_n = \sup F_n(x)\text{-}F(x)$ | KS can detect semantic changes by extracting embedding of the input data, which can be applied for more complex ML models with categorical feature levels. |
| Cramer-Von Mises (CVM) | $\sum_z^{z'} F(z)\text{-}F\text{-}ref''(z)^2 =$ | CVM could present better performance when dealing with higher shift moments by using the full joint of the samples. CVM tests are usually used for image systems instead of text or other categorical feature models. |
| Fishers Exact Test (FET) | $OR = \frac{\frac{N_1}{N^0}}{\frac{N^{ref}}{N_0^{ref}}}$ | FET test could monitor and detect the drift for feature model detection. $N_1$ and $N_o$ refer to the True and False value while j and ref present as test data and reference data respectively. |
| Maximum Mean Discrepancy (MMD) | $MMD(F, p, q) = \mu_p - \mu_q\ _F^2$ | MMD is a kernel-based method that measures the distance between different distributions with the mean embeddings among kernel Hilbert space F, which is used for image and text models. |
| Least-Squares Density Difference | $LSDD(p, q) = \int_x (p(x) - q(x)))^2 dx$ | LSDD could be used among image, text and different categorical feature models. Drifts can be detected by computing test statistics of LSDD value between two underlying distributions. |
| Classifier | $S_P := x_1,...,x\_n \sim P^n(X) \rightarrow S_Q := y_{-1},...,y\_n \sim Q^n(Y)$ | Drifts are monitored if the probabilities of unseen tests are significantly larger than unseen reference. The detector could be used for ML models among text, image, time-series and other kinds of categorical features. |
| Spot-the-diff(STF) | $logit(\hat{P}_T) = b_0 + b_J k(x, w_J)$ | STF test is a kind of extension of the Classifier which updates with a test window to enforce the performance on discriminating the feature generally. |

1) **Data Mismatch:** An incorrect training set can make the model deviate from the original goal. According to [155], Machine learning systems depend on data science, software engineering, and operations. However, the incorrect assumptions from these distinct perspectives make the system fail as result. Moreover, the incorrectness may be caused by human mistakes in research or systematic error during the training process.
2) **Old model problem:** The model has been used for too long, and it is no longer supported to be used in the current environment. In addition, the design purpose of the model deviates from the current demand, so it has been unable to provide users with good prediction results.
3) **Adversarial attack:** External risks always focus on the model's weakness and are always ready to attack the model through illegal methods. The purpose of adversarial attacks is to deceive the machine learning model as the malicious inputs and improve themselves to more effective offence to the system.
4) **Negative feedback:** There exist some datasets that affect the model in a negative way when it collects and automatically trains, ultimately causing the model to deviate from its original expectations.

In addition, model monitoring follow the necessary principles because monitoring itself is to serve the original model and improve its capabilities, rather than building a new model to solve a new problem. Therefore, researchers should consider the following three points when designing the model. (1) Monitoring and the goals of the machine learning model should be aligned, and the monitoring purpose is to serve the model. (2) The effect of the original system should be avoided without reducing the efficiency of the system. (3) Monitoring should be able to make changes as required by the model and make effective strategy adjustments in the face of different situations.

Based on the existing research, many monitor model experts put forward theories about the monitor models. The researchers try to make the monitoring model work better and help the system through different algorithms.

**Completely Same Monitoring:** A model that is identical to the original model is deployed, and how the original model works are checked by observing of the differences between the new model and the original model.

**Partly Same Monitoring:** Only monitoring is deployed to the key parts of the model. This monitoring system can exist in the original model, or it can be completely independent.

**Different Monitoring:** A new model is built with a different method, but the new model should be consistent with the original model's goals, to monitor the original model with the results.

**Data Monitoring:** It mainly depends on regression algorithms and classification algorithms. The aim is to randomly check that real data still exists in an acceptable range[156]. This range should neither be too strict about causing overfitting, nor too broad to cause underfitting.

**Prediction Monitoring:** The distribution of prediction results of the model is used to compare the statistical tests with the following values: maximum values, minimum values, median, mean, mode, ans standard deviation. If their differences are acceptable, then the monitoring is working well and the model is not in trouble. However, if the actual situation deviates from what we expected, the researcher should observe the difference and use it to re-evaluate the model[157].

Finally, it is worth paying attention to the risk of monitoring. In addition to the original external intrusions that can harm the monitoring model, the monitoring process effects can also negatively affect it. Although the researcher tries to use monitoring to keep the quantity of the model, there still exist some risks of monitoring, that could threaten our target.

**Hazard system:** Monitoring can affect the overall productivity of the model because it takes up system resources. Thus the entire model can be at risk of delays or freezes. According to [158], as time goes by, the model will decrease its fitting and forecasting ability.

**Complex:** Monitoring complicates the overall model environment and magnifies the model's butterfly effect. Therefore, the researcher should be more careful in adjusting parameters to ensure the consistency of the model.

## 11.3   Compliance

Machine learning models need to comply with industry-specific standards and government regulations, defining the basic requirements of deployed models. Failure of compliance may pose risks to models, and termination of the operation. Specifically, a series of studies have been conducted, to explore how machine learning can preserve compliance with relevant regulations. There are three mainly situations of compliance that need to be noticed.

1) The objective of deployed models is to ensure the compliance of specific applications. For example, ML can be exploited in financial compliance management [210, 211].

2) Deployed models are utilised in specific industries with industry specifications and standards.Bharadwaj [212] focuses on auditing of robot learning, to ensure the safety and compliance. Vasudevan et al. [213] propose a framework for the compliance of machine learning models with automotive industry standards.

3) Machine learning models need to comply with legal regulations issued by the local government. For instance, several studies discuss how machine learning can comply with the General Data Protection Regulation [204–206]. Kingston [204] suggests to focusing on compliance checklists and codes of conduct, risk assessments, automatic profiling, and breaches of security. Oliver et al. [205] and Goldsteen et al. [206] explore data anonymisation and minimisation in machine learning respectively.

## 11.4   System Logging

: In ML applications, logging is a method to surveillance ML pipeline and detect the healthy of the ML models in flexibly and convincing way. The logs that gathering by programs guide the next itanration of a ML system. It is a critical process for all ML applications because wrong logging decisions may be likely to cause the whole project to fail or result in unacceptable performance. The increasing need to deal with ML logging is receiving attention from both the ML and software engineering communities [95, 200, 202]. In general, ML logs consist of the following three components,

1) **logging for ML model**: The model logging statements include the ML model tracking information which consists of the model publisher, the model name and the version number. The next part of logging is model changes management information, and when models were deployed in the production environment. Moreover, the set of hyper-parameters used for a model is another important information to be logged if we want to reproduce the same result repeatedly set.

2) **Tracking ML system:** ML system logs can help with attribution anal- ysis and trace the root cause of existing defects. ML System logs include dataset tracking, monitoring accuracy at each iteration of the training, testing, final test results and different level of evaluation matrix. To create a unified view of ML systems, we need information from input data sources to model predictions and end up with a dashboard tvthe isualiseze overall system.

3) **Infrastructure logging:** Obtaining infrastructure level logs are the over-arching process of monitoring ML applications because a ML application can be launched on cloud-based services, containerised systems, micro-services or Internet of things edges infrastructure. This means obtaining logs relies on a combination process that aggregates information from both infrastructures providers and ML application creators.

Besides what information was inserted in the logging statements, log distribution is another feature that researchers are concerned about. Fu [207] describes logs granularity as quit difficult to decide in ML practice because wrong granularity can negatively impact system performance. during ML development, algorithm engineers and MLOps engineers are likely to insert too little or too much tracking information. In addition, choosing a proper log level is another challenge for a ML application. Guidance on determining the log level was proposed by Li [202]. A well-defined logging strategy can increase ML systems' observability and controllability. It is also helpful to detect the root cause of system failure and indicate how well it achieves the business goals.

## 11.5   Model Explanation

With the increasing prevalence and complexity of methods, data-specific biases, and growing numbers of concerns about the decision making of AI models from business stakeholders, Machine Learning (ML) explanation has been a significant challenge. Explainable artificial intelligence (XAI) techniques help answer essential questions about *how* and *why* complex AI systems make a decision. It also provides a means to address growing ethical and legal issues. As a result, AI researchers have identified XAI as a necessary feature of trustworthy AI, and explainability has experienced a recent surge in attention. However, despite the growing interest in XAI research and the demand for explainability across disparate domains, XAI still suffers from several limitations. This section provides an overview of the current state of XAI, as well as its advantages and disadvantages.

In DARPA's explainable AI (XAI) program [208], the psychological model of explanations is also investigated. An in-depth discussion of explanation along this line studying the problem from a social science perspective can be found in [209]. This paper focuses on DNN model explainability in the context of MLOps.

In MLOps, practitioners and developers utilise explanation tools to validate AI system implementation against project requirements, assist problem diagnostics in debugging, and identify problems in testing. Users and regulators may benefit from explanations to obtain a better knowledge of how AI systems function and to address issues and concerns regarding their behaviour and the auditability of the system.

The primary goals of XAI is to provide explainability to AI systems including the following[4]: (1) assisting human decision making and improving trust to a certain level; (2) providing transparency to the complex optimisation process; (3) providing information for model debugging; (4) enabling auditing and accountability.

There are two main approaches to achieving explainability:

1. **Post-hoc explanation techniques**: seek insight into a trained model to find links to predictions [94].
2. **Build-in explainable models**: intentionally guide model decisions to focus specific part of the input or relate to a specific "prototype" in reasoning to achieve end-to-end explainability [93];

### 11.5.1   Post-hoc explanation techniques

Post-hoc explanation techniques can be categorised as follows:

1) **Feature attribution methods**: compute the importance of each dimen- sion of an input data sample on the model output of this sample. Commonly used techniques include *Input Gradient* [216] and its improvements, such

---

as *Smoothgrad* [219], *Integrated Gradients* [218], and *Guided Backpropagation* [220].

Consider the activation function of a specific class *c* of a DNN-based model is argmax($f_c(x)$) for a given input sample *x*, the saliency map of $f_c$ with respect to *x*, denoted by $s_c(x)$ is therefore as below:

$$s_c(x) = \frac{\partial f_c(x)}{\partial x}$$

If *x* is an image, $s_c(x)$ shows how the prediction result will change if a pixel in the image has a tiny change. The degree of difference in the prediction result indicates the importance of a pixel.

Smoothgrad improves Input Gradient by adding Gaussian noise to the input *x* in attempt to smooth out the fluctuation of the derivatives of $f_c$ so that the saliency map may reflect the overall sensitivity better.

The Integrated Gradients method further attributes the prediction results with respect to a baseline $x_0$ by interpolating the change between $x_0$ and *x*. The saliency map is formed by aggregate $s_c(x)$ along the interpolating path as below:

$$IG(x) = (x - x_0) \int_{\alpha=0}^{1} \frac{\partial f_c(x_0 + \alpha(x - x_0))}{\partial x} d\alpha$$

in which $\alpha$ is the interpolation step. Integrated Gradients can also be applied to graph neural networks to explain node classification or link prediction results [223].

Guided Backpropagation builds on a deconvolution approach to attribute the prediction result to the input. It handles the nonlinear RELU layer differently in backpropagation by setting negative gradient values to zero. This prevents negative gradient values in higher layers from flowing backwards and helps better visualise the higher layer activation.

Guided Grad-CAM [221] computes the gradient of predictions for class *c* with respect to the feature map of a convolution layer rather than the input *x*. It uses Guided Backpropagation to obtain pixel saliency map.

2) **Concept alignment methods**: try to identify the relationship between individual hidden units and a set of human-understandable concepts [217]. The concept is given using examples in [222] and then its representation is obtained as a vector in the space of activations of a DNN layer. The vector can then be used to compute the sensitivity of the input towards the concept.

3) **Local explanation methods**: such as LIME [225] approximate the DNN model locally with interpretable models. These interpretable models behave faithfully to the DNN model for local data points of concern.

4) **Perturbation-based methods**: compute feature importance through observing the model output change by perturbing input features. This may help to understand model behaviours under external constraints. Shapley-value calculation [224, 227] is often used to achieve this purpose [226].

5) **Training data ranking based methods**: [228] rank the training samples that influence to the prediction of a DNN model on a test sample. Such methods may play an important role in MLOps by identifying problems in training data.

### 11.5.2    Build-in explainable models

Post-hoc approaches have been criticised as providing explanations that do not faithfully reflect what the original models do [229]. They do not improve the trust of models among users [231] even though they are practically used as debugging tools among developers. However, the observations of unstable saliency maps [230] indicate that build-in explainable models are likely to be an increasingly important approach for developing trustworthy ML systems.

**(author?)** [229] argues that incorporating interpretability constraints in model development will provide many benefits for high stake applications. For high-stake applications, it suggests that a build-in explainable model should be used wherever the model can achieve a comparable performance comparing to a black-box model. However, a build-in explainable model, such as decision trees, linear regression and rule-based models can only deal with problems with relatively low complexity.

In recent years, new approaches have shown certain success in injecting reasoning processes into complex models. **(author?)** [232] introduce a prototype layer in a DNN model to find parts of training images that represent a specific class. These parts are used as a prototype for classification. The optimisation process uses the similarity between prototypes and parts in the training images to minimise the classification error. When the model predicts a test image, it finds parts in the image that are similar to these prototypes for decision making.

[229] further argues that a set of DNN models is likely to exist that achieve similar performance on fitting a dataset given the large parameter space of DNN models. Finding such a build-in explainable model is highly possible with a network architecture that easily explains its own reasoning process. Another approach to implement build-in explainable models is *mimic learning* [233], which distils soft labels from deep networks to learn interpretable models such as Gradient Boosting Trees and derive strong prediction rules.

## 12  summary

In this paper, we surveyed comprehensively machine learning technologies with respect to the whole machine learning process. We represent a machine learning process model, which is based on viewing machine learning as machine learning operations (MLOps). Our machine learning model doesn't just focus on only one specific machine learning branch or part of the machine learning life cycle. Instead, we cover the whole process, which aims to bring a global view of the whole picture of ML and linkages between the activities and concepts of each step. For each step of our model, we illustrated the key concepts and activities by collecting representative technologies, algorithms and applications. This survey paper can serve as a quick reference for researchers and/or someone new to ML to get a clear picture of ML.